# Malware Classification from Memory Dumps Using Machine Learning, Transformers, and Large Language Models

Areej Dweib
*Department of Natural, Engineering and Technology Sciences, Faculty of Graduate Studies*
*Arab American University*
Ramallah, Palestine
a.dweib@student.aaup.edu

Montaser Tanina
*Department of Natural, Engineering and Technology Sciences, Faculty of Graduate Studies*
*Arab American University*
Ramallah, Palestine
m.tanina@student.aaup.edu

Shehab Alawi
*Department of Natural, Engineering and Technology Sciences, Faculty of Graduate Studies*
*Arab American University*
Ramallah, Palestine
s.alawi@student.aaup.edu

Mohammad Dyab
*Department of Natural, Engineering and Technology Sciences, Faculty of Graduate Studies*
*Arab American University*
Ramallah, Palestine
m.dyab@student.aaup.edu

Huthaifa I. Ashqar
*Civil Engineering Department*
*Arab American University*
Jenin, Palestine
huthaifa.ashqar@aaup.edu

***Abstract*—This study investigates the performance of various classification models for a malware classification task using different feature sets and data configurations. Six models—Logistic Regression, K-Nearest Neighbors (KNN), Support Vector Machines (SVM), Decision Trees, Random Forest (RF), and Extreme Gradient Boosting (XGB)—were evaluated alongside two deep learning models, Recurrent Neural Networks (RNN) and Transformers, as well as the Gemini zero-shot and few-shot learning methods. Four feature sets were tested: All Features, Literature Review Features, the Top 45 Features from RF, and Down-Sampled with Top 45 Features. XGB achieved the highest accuracy of 87.42% using the Top 45 Features, outperforming all other models. RF followed closely with 87.23% accuracy on the same feature set. In contrast, deep learning models underperformed, with RNN achieving 66.71% accuracy and Transformers reaching 71.59%. Down-sampling reduced performance across all models, with XGB dropping to 81.31%. Gemini's zero-shot and few-shot learning approaches showed the lowest performance, with accuracies of 40.65% and 48.65%, respectively. The results highlight the importance of feature selection in improving model performance while reducing computational complexity. Traditional models like XGB and RF demonstrated superior performance, while deep learning and few-shot methods struggled to match their accuracy. This study underscores the effectiveness of traditional machine learning models for structured datasets and provides a foundation for future research into hybrid approaches and larger datasets.**

## I. Introduction

Malicious software, commonly known as malware, poses a cybersecurity risk on a scale. These harmful programs are specifically created to disrupt, harm or gain entry, into computer systems. As technology advances the complexity of malware also increases, making detection methods based on signatures effective. This has led to the adoption of approaches like machine learning and memory forensics to identify and categorize malware more efficiently [1].

The threat of malware is significant in the realm of cyber-security with the AV TEST Institute recording over 450,000 programs every day [2]. In 2022, there were 5.5 billion global malware attacks—a 2% uptick from the year—contributing to 40% of data breaches in 2023 marking a rise of 30% from 2022 [3]. The education sector witnessed a surge of 157% in malware incidents between 2021 and 2022 while the healthcare industry experienced a decrease of 15%. Notably there were increases in malware attacks within the retail and finance sectors with year over year spikes of 50% and 86% [4]. It is imperative to classify malware into groups or families, for various reasons. First, this allows cybersecurity experts to grasp the behavior and traits of types of malwares which aids in developing defense strategies [5]. Secondly, it assists in identifying and responding to malware incidents thereby reducing harm. Furthermore, effective classification supports efforts in threat intelligence by helping predict and minimize attacks [6].

The main goal of this study is to create a machine learning model that can categorize malware samples into families or groups based on evidence extracted from memory dumps. Memory dumps provide insights into the state of a system at a moment including active processes, loaded modules and network connections. By examining these clues our aim is to develop an adaptable model that enhances the accuracy of detecting and classifying malware. Additionally, this research seeks to compare the result of four traditional machine learning algorithms with models, like Transformer and Multimodal Large Language Models (MLLMs). In order to accomplish the specified goal this study aims to explore three research questions. 1) What are the optimal classical machine learning algorithms (Decision Tree, Random Forest, Support Vector Machine, Recurrent Neural Network) for categorizing malware? 2) How do Transformer models and MLLMs perform compared to machine learning algorithms in malware classification? 3) What are the best set of features that can be used for this task?

This study leverages traces from memory dumps, which are often underutilized to enhance the identification and sorting of malware. Our objective is to offer a tool that does not only enhance cybersecurity but also contributes to the wider realm of cybersecurity research. Through a comparison of classical machine learning algorithms with Transformer and MLLMs, this research strives to pinpoint the efficient methods, for precise and effective malware classification.

## II. Related Work

Several studies have endeavored to enhance the precision and efficacy of malware detection and categorization by leveraging forensic artifacts extracted from memory dumps. Memory forensics, coupled with feature engineering, has emerged as a pivotal approach in malware classification. Notably, the VolMemLyzer tool showcased the potential of feature engineering by achieving a 93% true positive rate using decision trees and random forests on memory dump features [7, 8]. One-class classifier models have proven to be effective in memory dump malware detection, with an enhanced OCSVM model integrated with Principal Component Analysis (PCA) achieving a remarkable 99.4% accuracy on the MALMEMANALYSIS2022 dataset [9, 10].

The application of computer vision techniques in memory forensics-based malware detection has yielded promising results. A study achieved a 97.01% accuracy rate employing a combination of contrast-limited adaptive histogram equalization and wavelet transform with machine learning classifiers such as support vector machines (SVM), random forests, decision trees, and XGBoost [11]. Additionally, the effectiveness of K-nearest neighbors (KNN) in identifying memory malware payloads underscores the significance of various machine learning algorithms for real-time memory malware identification [12].

Decision tree algorithms based on memory features engineering have demonstrated exceptional accuracy, with one study reporting a 99.982% accuracy rate and a low false positive rate [13]. Furthermore, lightweight multiclass malware detection methods tailored for resource-constrained IoT devices have been developed, with models achieving high accuracy on obfuscated memory malware in smart city applications [14]. A study compared methods and models to find the approach for classifying malware emphasizing the importance of thorough evaluation [15]. By combining CNN and LSTM networks another research project achieved accuracy by con- verting files into grayscale images [16]. Convolutional Fuzzy Neural Networks, along with feature fusion and the Taguchi method tuned parameter combinations to boost accuracy [17]. Various classifiers like SVM, KNN and decision trees were employed to improve performance [18]. Genetic Algorithm CNN techniques proved effective, in malware classification with accuracy while an enhanced KNN model outperformed strategy [19]. Using normalized input images reduced training time without compromising accuracy [20].

In addition to memory forensics, cloud-based machine learning workflows have been scrutinized for their scalability and efficiency. Comparative evaluations of Amazon Web Ser- vices (AWS), Google Cloud Platform (GCP), and Microsoft Azure underscore the importance of leveraging cloud plat- forms for building robust machine learning workflows for tasks such as image classification, which can be extended to malware classification [21]. However, Deep learning algorithms have garnered significant attention for malware classification tasks. Studies have proposed CNN- based algorithms to identify and categorize malware families based on color-based images, achieving superior accuracy compared to traditional models [22]. Other research endeavors have focused on converting malware binaries to grayscale images and employing CNNs for classification, showcasing the effectiveness of visual categorization for malware detection [23].

The literature reflects a trend toward employing deep learning, hybrid models, and feature engineering to classify malware based on forensic artifacts extracted from memory dumps. These advancements aim to address the challenges posed by modern malware and enhance the accuracy and efficiency of malware detection and classification across diverse computing environments. Our contributions aim to advance the field of malware classification by developing a robust, scalable, and accurate model that leverages forensic artifacts from memory dumps, integrates advanced feature engineering and hybrid machine learning techniques, and addresses the practical needs of modern computing environments. By comparing traditional classifiers with cutting-edge Transformer and MLLM architectures, we provide comprehensive insights into the most effective approaches for malware classification and what are the best set of features that can be used for this task.

## III. Methodology

### A. Dataset Description

This method replicates scenarios that an average user might face when confronted with a malware attack [6]. We used MalMem2022 dataset. The dataset includes 58 features across 58,596 rows. The dataset is balanced only for each main malware class (50%) with about 29,298 observations as shown in Fig. 1.

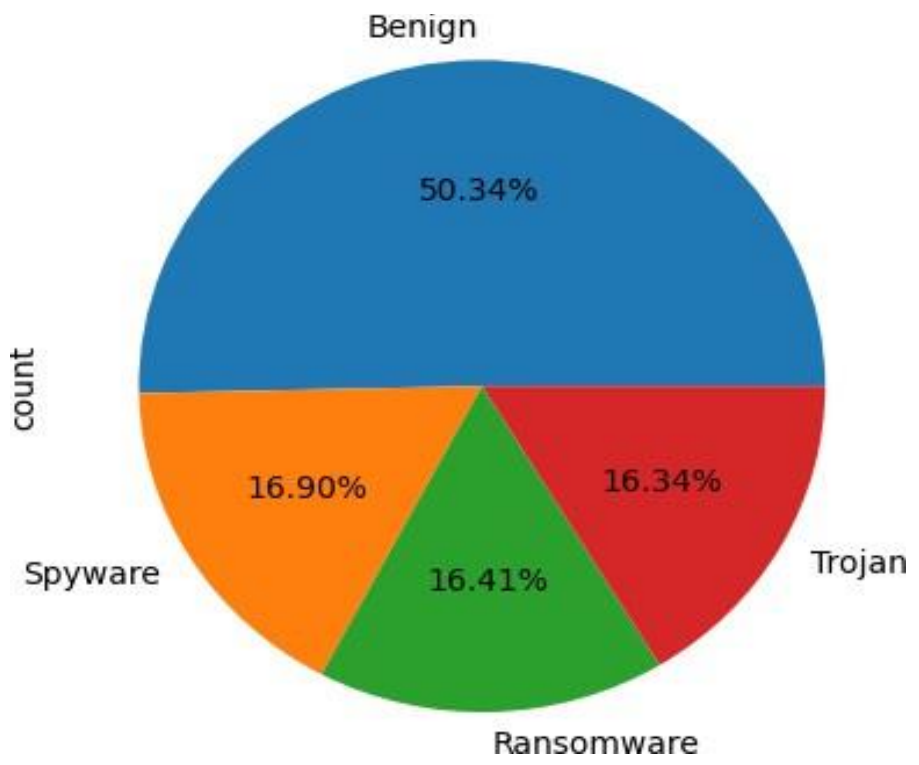

Fig. 1. The percentage of each class in the dataset.

For feature selection, we used four different feature sets to evaluate the performance of the methodology. Each feature set represents a unique selection of attributes, which were carefully chosen to test and refine the model's effectiveness. The following feature sets were utilized. For **All Features**, this set includes all available features in the dataset without any filtering or prioritization. It provides a comprehensive input for the model, ensuring no potential information is excluded. However, this approach can introduce noise or redundant data, which may affect the performance. For **Literature Review Features**, these features were selected based on prior studies and domain expertise. By leveraging knowledge from existing research, this set focuses on attributes that are most likely to be relevant, reducing noise and improving interpretability.

For **Top 45 Features from RF**: Random Forest was used to rank the features by importance, and the top 45 were chosen as shown in Fig. 2. This selection ensures the model uses only the most influential features, potentially improving accuracy and efficiency by reducing computational complexity. For **Down-Sampled with Top 45 Features**, to address class imbalance in the dataset, we combined down-sampling with the top 45 features. This approach balances the data distribution while maintaining the most critical attributes, improving the model's ability to generalize and perform well across all classes.

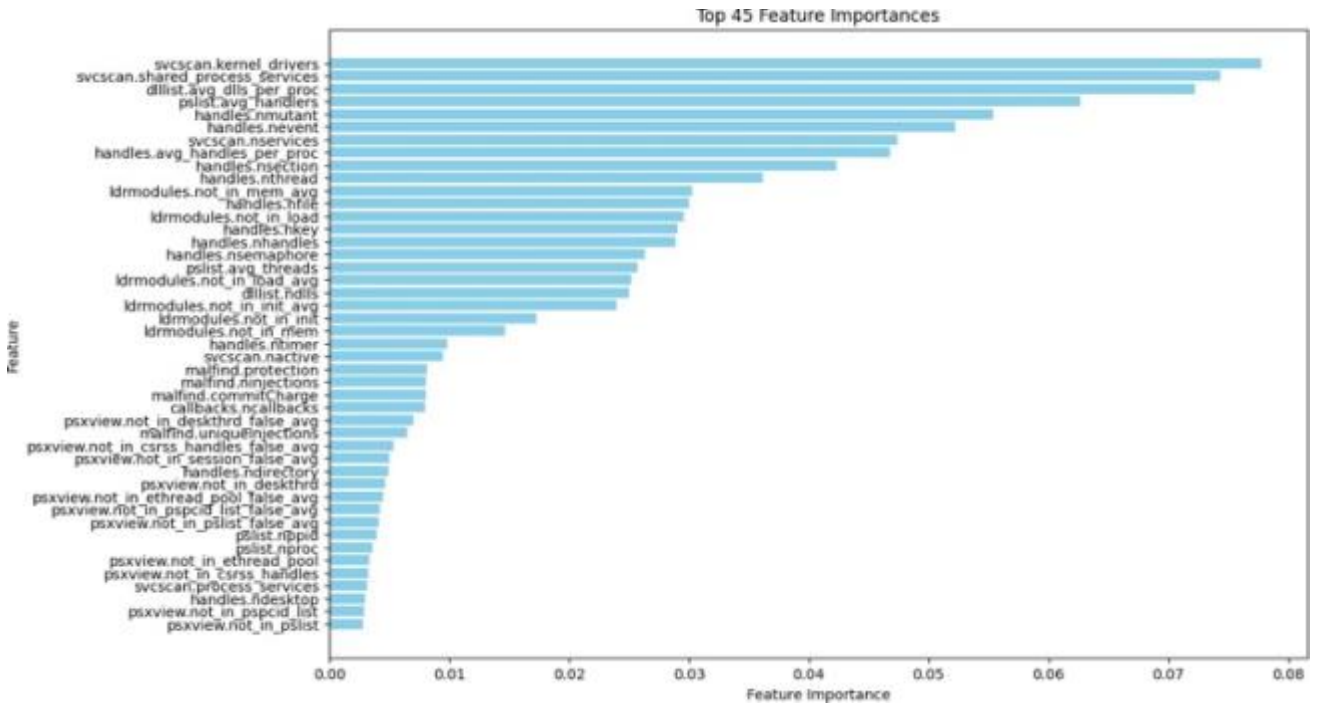

Fig. 2. Top 45 features and their weight.

Feature selection plays a crucial role in machine learning and predictive modeling. It reduces the dimensionality of the data, enhances model performance, and decreases the risk of overfitting. By comparing the outcomes of different feature sets, we ensured the methodology is robust and optimized for the task. Moreover, selecting features based on prior research and statistical importance adds reliability and interpretability to the results, aligning the methodology with both scientific rigor and practical application.

### *B. Classification Algorithms*

We compare between some of techniques that uses in classification the malware, we begin with the classic machine learning algorithms and train the model on the dataset, the algorithms that we work on there are: Logistic Regression, K-Nearest Neighbors, Support Vector machine, Decision Tree, Random Forest and Recurrent Neural Network (RNN). After that we use transformer and Gemini as LLM models. All this work for develop a machine learning model to classify malware into families based on memory dump analysis. Traditional approaches, such as signature-based detection, often fail to identify malware that does not have predefined signatures (Zero-day malware).

#### *1) Classical Classification Algorithms*

**Logistic Regression** is used technique, for classifying malware in categories because of its simplicity, ease of understanding and effectiveness. Even though it's a model it can handle classes by using methods like one vs rest (OvR) or SoftMax regression. This method is especially useful in situations where quick predictions are needed in time since it does not require as much computational power as more complex models [24].

**K-Nearest Neighbors (KNN)** is a technique, for classifying malware across categories because of its straightforward approach, efficiency and adaptability. It works by comparing data points to known examples and assigning them a class based on most of their neighbors. This method is particularly advantageous in malware detection as it does not make assumptions about data distribution allowing it to be versatile and adaptable to types of malware patterns.

**Decision Trees** are often chosen for categorizing malware into classes because of their simplicity, clarity and ability to handle types of data. This approach involves dividing the data into groups based on characteristics creating a decision tree structure. A key advantage of using Decision Trees in malware analysis is their transparency; each decision taken by the model can be easily tracked back through the tree enabling security experts to understand why certain classifications were made. Decision Trees can manage relationships between features without needing data preparation or adjustments [25].

**Random Forest** is considered effective for categorizing types of malwares because of its reliability, precision and capability to manage extensive and intricate datasets. By utilizing decision trees in an approach, it avoids overfitting issues and improves its ability to predict outcomes for new data [26]. This method excels at handling a range of features. Provides valuable insights into the significance of each feature making it easier to understand the results. It is not easily affected by noise. It can maintain accuracy levels even when dealing with incomplete or corrupted data. Despite its complexity Random Forest can be scaled up efficiently.

**Recurrent Neural Networks (RNNs)** are highly effective for multiclass malware classification due to their ability to handle sequential data and capture temporal dependencies, which are crucial for analyzing malware behaviors over time. They process input sequences of varying lengths and adapt to different malware signatures, making them ideal for understanding the evolution of malware activity.

**Support Vector Machine (SVM)** is known for its ability to effectively classify data especially when dealing with large datasets and aiming for accuracy. What sets SVM apart is its interpretability, which has contributed to its use across fields.

We used these six algorithms and apply them in the dataset to find the best one in the accuracy, precision, recall, F1 score, all of algorithms that use with the default parameter. in the next section we will talk about transformer.

#### *2) Transformers Classification Algorithms*

Transformers such as TabTransformer model, have greatly improved AI abilities in tasks involving natural language processing such as categorizing text, analyzing emotions and detecting areas like malware. Unlike RNNs or CNNs Transformers use self-attention techniques to understand the connections between words in a sentence or context which helps them grasp the subtleties and meanings of language better. The TabTransformer is a tool for organizing data, like a sophisticated spreadsheet. By utilizing self-attention, it examines the connections between features to determine their impact on each other, generates embeddings for data comprehension and adapts effectively to handle sizable datasets. Drawing inspiration from natural language processing techniques, this method results in improved model accuracy for classification tasks. A comparison between RNN and TabTransformer is shown in TABLE I.

TABLE I
PARAMETERS OF DEEP LEARNING ALGORITHM (RNN & TRANSFORMERS)

| Parameter | RNN | Transformer |
|---|---|---|
| Units | 64 | 128 |
| Return Sequences | True | False |
| Loss | Sparse_categorical_crossentropy | nn.CrossEntropyLoss() |
| Optimizer | Adam | Adam |
| Epochs | 6 | 3 |
| Batch Size | 128 | 512 |

*3) MLLM Models*

For this part of the study, we used **Gemini** to classify malware into two levels [27, 28]. For **Zero-Shot Learning**, we provided the Gemini model with the dataset without specifying the target labels and asked it to classify the data into four categories: Benign, Trojan, Ransomware, and Spyware. The model relied solely on its pre-existing knowledge to perform the classification without any prior training on this specific dataset. We used the following prompt for zero-shot learning: *"You are a cybersecurity expert, and you have a dataset for memory dumps. Your task is to categorize the data into 4 categories: ransomware, benign, spyware, and Trojan. If you understand, respond with 'yes' so I can send you the samples."*

**For Few-Shot Learning**, before asking Gemini to classify the dataset, we provided it with a small training set of 8-labeled observations, representing examples from each of the four categories. This step allowed the model to learn from these examples before performing the classification task. We used the following extended prompt for few-shot learning: *"You are a cybersecurity expert, and you have a dataset for memory dumps. Your task is to categorize the data into 4 categories: ransomware, benign, spyware, and Trojan. This is an example of each category. If you understand, respond with 'yes' so I can send you the data to categorize."*

## IV. Evaluation Results

This section describes the evaluation results achieved by ap- plying each one of the models training over the MalMem2022 dataset. TABLE II compares the performance of various classification models using different feature sets. These models are evaluated using metrics like accuracy (A), F1 score (F1), precision (P), and recall (R). The results demonstrate the impact of feature selection and dataset size on model performance, as well as the effectiveness of traditional machine learning models compared to Gemini's zero-shot and few-shot learning approaches.

Using the full feature set, XGB achieved the highest accuracy (87.33%), closely followed by RF at 87.05%. Decision Tree also performed well, reaching 84.53% accuracy. Logistic Regression lagged behind, achieving the lowest accuracy in this group at 74.65%. Overall, XGB and RF demonstrated their ability to effectively utilize the full feature set, making them the top-performing models for this configuration. Using features derived from literature slightly reduced the performance of all models compared to the "All Features" set. XGB remained the best-performing model, though its accuracy dropped to 84.77%. Random Forest and Decision Tree experienced similar decreases, with their accuracies dropping to 84.29% and 82.01%, respectively. These results suggest that the literature-based feature set, while useful, may not capture as much predictive power as the full dataset.

Feature selection, specifically using the top 45 features determined by RF importance, slightly improved the performance of the best models. XGB achieved the highest accuracy in this category at 87.42%, while RF followed closely at 87.23%. Other models, such as KNN and SVM, maintained similar performance to the all-features set. However, deep learning models like RNN and Transformer underperformed, with accuracies of 66.71% and 71.59%, respectively. This indicates that traditional models handle this specific dataset and feature set better than deep learning approaches. Down-sampling the dataset resulted in a noticeable performance drop for all models. XGB remained the best model, achieving 81.31% accuracy, while RF followed at 80.52%. However, these scores were about 6% lower than when the full dataset was used. Logistic Regression struggled the most, achieving only 61.50% accuracy. These results suggest that while down-sampling may help balance the dataset, it also leads to significant information loss, negatively impacting model performance.

The Gemini model performed poorly compared to traditional machine learning models. In the zero-shot learning scenario, where no prior training data was provided, the model achieved the lowest accuracy across all methods (40.65%). Few-shot learning, where a small labeled dataset was introduced, slightly improved accuracy to 48.65%. Despite this improvement, Gemini's performance remained far below that of the traditional models.

Feature selection proved beneficial, as reducing the dataset to the top 45 features slightly enhanced the performance of top models like XGB and RF. These models consistently outperformed others across all feature sets, making them the most reliable for this task. However, down-sampling the dataset negatively affected performance, likely due to the loss of critical information. Gemini's low performance in both zero-shot and few-shot scenarios underscores the importance of sufficient training data for effective classification.

TABLE II
Comparison of models and feature sets

| Features Set | Model | A (%) | F1 (%) | P (%) | R (%) |
|---|---|---|---|---|---|
| **All Features** | Logistic Regression | 74.65 | 74.53 | 75.49 | 74.65 |
| | K-Nearest Neighbors | 80.50 | 80.51 | 80.64 | 80.50 |
| | Support Vector | 76.59 | 76.28 | 77.67 | 76.59 |
| | Decision Tree | 84.53 | 84.54 | 84.54 | 84.53 |
| | Random Forest | 87.05 | 87.05 | 87.05 | 87.05 |
| | Extreme Gradient Boosting | 87.33 | 87.31 | 87.30 | 87.33 |
| **Literature Review Features** | Logistic Regression | 72.22 | 71.89 | 73.08 | 72.22 |
| | K-Nearest Neighbors | 79.69 | 79.68 | 79.79 | 79.69 |
| | Support Vector | 75.15 | 74.61 | 76.91 | 75.15 |
| | Decision Tree | 82.01 | 82.00 | 82.00 | 82.01 |
| | Random Forest | 84.29 | 84.29 | 84.29 | 84.29 |
| | Extreme Gradient Boosting | 84.77 | 84.77 | 84.77 | 84.77 |
| **Top 45 Features from RF** | Logistic Regression | 74.41 | 74.26 | 75.28 | 74.41 |
| | K-Nearest Neighbors | 80.47 | 80.47 | 80.60 | 80.47 |
| | Support Vector | 76.76 | 76.44 | 77.82 | 76.76 |
| | Decision Tree | 84.44 | 84.45 | 84.47 | 84.44 |
| | Random Forest | 87.23 | 87.22 | 87.22 | 87.23 |
| | Extreme Gradient Boosting | 87.42 | 87.39 | 87.39 | 87.42 |
| | RNN | 66.71 | 58.52 | 55.77 | 66.71 |
| | Transformer | 71.59 | 68.26 | 71.90 | 71.59 |
| **Down-Sampled with Top 45 Features** | Logistic Regression | 61.50 | 61.31 | 62.49 | 61.50 |
| | K-Nearest Neighbors | 71.09 | 71.06 | 71.30 | 71.09 |
| | Support Vector | 65.88 | 65.50 | 66.98 | 65.88 |
| | Decision Tree | 76.33 | 76.34 | 76.35 | 76.33 |
| | Random Forest | 80.52 | 80.52 | 80.52 | 80.52 |
| | Extreme Gradient Boosting | 81.31 | 81.29 | 81.29 | 81.31 |
| **Genimi** | Zero Shot | 40.65 | 46.13 | 51.40 | 43.86 |
| | Few Shot | 48.65 | 50.00 | 52.74 | 49.22 |

## V. Conclusion

This study evaluated the performance of various models across different feature sets for the classification task, providing valuable insights into the importance of feature selection, dataset size, and model choice. The findings highlight that traditional machine learning models, particularly XGB and RF, consistently delivered the highest accuracy and overall performance metrics across all configurations. These models excelled at leveraging both the full feature set and the reduced set of the top 45 features, demonstrating their robustness and adaptability.

Feature selection emerged as a critical factor for improving model efficiency and accuracy. Using the top 45 features identified by RF slightly enhanced the performance of the best models, indicating that selecting relevant features can reduce computational complexity without compromising predictive power. However, down-sampling the dataset led to a noticeable decline in performance for all models, emphasizing the importance of maintaining sufficient data to capture key patterns and relationships.

The deep learning models, RNN and Transformer, underperformed compared to traditional models. This suggests that for this dataset and task, simpler algorithms may be better suited due to their ability to handle structured features effectively without the need for extensive computational resources. However, the Gemini model, employing zero-shot and few-shot learning, achieved significantly lower performance than traditional machine learning approaches. While these methods are promising for tasks with limited labeled data, their results in this study underscore the need for more robust training data and feature representation to improve classification accuracy in such scenarios.

Results showed that the combination of feature selection and traditional machine learning models, particularly XGB and RF, proves to be the most effective approach for this classification task. Future work could focus on enhancing Gemini's capabilities through advanced fine-tuning and exploring hybrid models that integrate deep learning and traditional algorithms to leverage the strengths of both approaches. Additionally, testing these methods on larger and more diverse datasets could provide deeper insights and broader applicability of the findings..